\documentclass[lettersize,journal]{IEEEtran}
\usepackage{amsmath,amsfonts}
\usepackage{algorithmic}
\usepackage{algorithm}
\usepackage{array}
\usepackage[caption=false,font=normalsize,labelfont=sf,textfont=sf]{subfig}
\usepackage{textcomp}
\usepackage{stfloats}
\usepackage{url}
\usepackage{verbatim}
\usepackage{graphicx}
\usepackage{cite}
\usepackage{xcolor}
\usepackage{multirow}
\usepackage{amsfonts}
\usepackage[colorlinks,bookmarksopen,bookmarksnumbered, citecolor=blue, linkcolor=red, urlcolor=blue]{hyperref}
\usepackage{amsmath}
\usepackage{amssymb}

\begin{document}

\title{Federated Progressive Self-Distillation with Logits Calibration for Personalized IIoT Edge Intelligence}

\author{\IEEEauthorblockN{Yingchao Wang\IEEEauthorrefmark{1} and Wenqi Niu\IEEEauthorrefmark{2}}

\IEEEauthorblockA{\IEEEauthorrefmark{1}School of Cyberspace Science and Technology, Beijing Institute of Technology, Beijing, China.}

\IEEEauthorblockA{\IEEEauthorrefmark{2}School of Electrical and Information Engineering, North Minzu University, Ningxia, Yinchuan, China.}
\thanks{Corresponding author: Yingchao Wang (email: yingchaowang@bit.edu.cn).}}




\markboth {}
{Shell \MakeLowercase{\textit{et al.}}: A Sample Article Using IEEEtran.cls for IEEE Journals}


\maketitle
\begin{abstract} Personalized Federated Learning (PFL) focuses on tailoring models to individual IIoT clients in federated learning by addressing data heterogeneity and diverse user needs. Although existing studies have proposed effective PFL solutions from various perspectives, they overlook the issue of forgetting both historical personalized knowledge and global generalized knowledge during local training on clients. Therefore, this study proposes a novel PFL method, Federated Progressive Self-Distillation (FedPSD), based on logits calibration and progressive self-distillation. We analyze the impact mechanism of client data distribution characteristics on personalized and global knowledge forgetting. To address the issue of global knowledge forgetting, we propose a logits calibration approach for the local training loss and design a progressive self-distillation strategy to facilitate the gradual inheritance of global knowledge, where the model outputs from the previous epoch serve as virtual teachers to guide the training of subsequent epochs. Moreover, to address personalized knowledge forgetting, we construct calibrated fusion labels by integrating historical personalized model outputs, which are then used as teacher model outputs to guide the initial epoch of local self-distillation, enabling rapid recall of personalized knowledge. Extensive experiments under various data heterogeneity scenarios demonstrate the effectiveness and superiority of the proposed FedPSD method.
\end{abstract}


\section{Introduction}
\label{Section I}
\IEEEPARstart{T}{he} Industrial Internet of Things (IIoT) has emerged as a critical enabler for future industries such as intelligent manufacturing, smart transportation, and smart healthcare, where interconnected devices generate and process vast amounts of data at the network edge \cite{yang2018internet, hu2024industrial}. To enable real-time distributed intelligent IIoT services and applications while ensuring data security and privacy, edge computing and federated learning (FL) have been extensively employed to support Artificial Intelligence (AI) driven IIoT applications \cite{yang2022cloud, nguyen2021federated, wang2024end}. Edge computing, on the one hand, brings computational resources closer to the data source, enabling localized data processing that reduces latency and improves response times in critical IIoT tasks, such as predictive maintenance and quality inspection. On the other hand, FL enables collaborative model training across distributed edge clients without the need to share raw data. Instead, only model updates are exchanged between devices and central servers, ensuring that sensitive data remains localized and secure.

However, traditional FL algorithms struggle to maintain robust performance in highly heterogeneous IIoT edge environments and lack personalized solutions tailored to each edge node's unique data characteristics. A fundamental issue lies in the Non-Independent and Identically Distributed (Non-IID) nature of data across different IIoT edge clients. In real-world IIoT systems, each edge device collects data influenced by its unique operational context, sensor types, geographic location, and usage patterns, resulting in significant variations in data distributions across clients. In such scenarios, the local models on each edge client tend to adapt to their specific data distributions during the FL process, leading to inconsistent parameter update directions during global model aggregation. This inconsistency negatively impacts the convergence and accuracy of the global model, potentially preventing the global model from effectively generalizing to the data of any particular client. 

Therefore, most existing FL studies \cite{li2020federated} aim to address the performance issues of the global model under Non-IID datasets, thereby enhancing the performance on local data. For example, FedProx \cite{li2020federated} introduced a regularization term in the local sub-optimization problem of FedAvg to limit the parameter distance between the local model and the global model in the parameter space. Although the performance of the global model has been effectively improved, significant differences in resource conditions, data distributions, and task objectives across clients make it challenging for a single global model to fully meet the specific needs of each IIoT client, leading to suboptimal performance in personalized IIoT tasks. This limitation may result in certain IIoT clients benefiting only marginally from FL, in some cases, the global model's performance may even be inferior to that of their locally trained models. This could reduce these IIoT clients' motivation to participate FL, ultimately affecting the overall effectiveness and scalability of FL in real-world IIoT deployments.

To address this issue, recent Personalized Federated Learning (PFL) \cite{tan2022towards} approaches have been introduced to develop personalized models specifically tailored to each client directly. From the perspective of knowledge transfer, the key to PFL is to achieve the effective integration and balance of global knowledge and local personalized knowledge, where the global knowledge comes from the global model in the aggregation and the local personalized knowledge comes from the local training \cite{yao2024rethinking}. Some studies \cite{lee2022preservation, lu2023federated, he2022learning, yao2023f} have recognized the importance of preventing global knowledge forgetting during local client training. For example, FedNTD \cite{lee2022preservation}, FedSSD \cite{he2022learning}, FedLMD \cite{lu2023federated} and FedGKD \cite{yao2023f} implemented knowledge distillation (KD)\cite{hinton2015distilling} technique on the local client to preserve the global perspective by distilling knowledge from the global model. However, retaining global models and performing distillation incurs non-negligible additional computational and storage overhead, which poses a significant burden for resource-constrained IIoT edge clients. On the other hand, the global model lacks sufficient confidence in the local data, which may potentially introduce erroneous knowledge into the local updates.

Moreover, these studies ignore the phenomenon of local personalized knowledge forgetting \cite{huang2022learn, jin2022personalized, he2022learning,  yao2024rethinking, su2025dkd}, which primarily arises from the tendency of the client's personalized parameters to be overshadowed or diminished by the ``averaging'' effect of the global model. Especially in real IIoT FL environments, edge clients may be unable to participate in every training round regularly due to unstable network connections, limited device resources, and strict local privacy policies. This non-uniform participation of clients can exacerbate the issue of local personalized knowledge forgetting. For example, when client $k$ has been absent from training for an extended period, the global model is more likely to diverge significantly from client $ k $'s personalized characteristics. Although local training can relearn personalized knowledge, the absence of historical personalized knowledge not only slows down the convergence of the local model but also hinders the model from gaining diverse insights into local data across different historical training stages. This, in turn, affects the model’s ability to effectively personalize.  

To simultaneously mitigate the forgetting of global knowledge and historical personalized knowledge during local training on resource-constrained IIoT clients, we propose a novel PFL framework Federated Progressive Self-Distillation (FedPSD) that incorporates time-dimensional self-distillation and knowledge calibration techniques. Specifically, to address the issue of global knowledge forgetting, on one hand, FedPSD dynamically calibrates the output logits during local training based on the data distribution of each IIoT client, mitigating the loss of global knowledge. On the other hand, FedPSD employs a progressive self-distillation strategy, where the model output from the previous epoch serves as a virtual teacher to guide the training of the next epoch, facilitating the gradual inheritance of global knowledge. Additionally, to tackle the problem of personalized knowledge forgetting, FedPSD constructs soft labels that integrate outputs from historical personalized models. These soft labels are used to guide the self-distillation process in the initial epoch, enabling rapid recall of personalized knowledge. Extensive experiments demonstrate that FedPSD significantly improves model personalization and overall performance without significantly increasing storage and computational overhead.

The main contributions of this study can be summarized as follows.
\begin{itemize}
	\item We theoretically demonstrate that the alternating process of aggregation and local training in FL leads to continual forgetting of both global and personalized knowledge.
	\item We propose a novel progressive self-distillation mechanism and a dynamic logits calibration technique in PFL to address the global knowledge forgetting challenge posed by Non-IID data in IIoT edge environments.
	\item We introduce calibrated fusion soft labels that incorporate clients' historical personalized knowledge to guide the self-distillation process of local models, enabling a rapid review of the historical personalized knowledge.
\end{itemize}

The rest of this paper is organized as follows. Related studies are reviewed in Section \ref{Section II}. Section \ref{Section III} presents the knowledge-forgetting issues and motivation of this study. Section \ref{Section IV}  demonstrates the details of FedPSD. Section \ref{Section V} shows the experiments in different Non-IID settings on MNIST, CIFAR-10, and CIFAR-100 datasets. Section \ref{Section VI} delves into the discussion. Finally, this paper is concluded in Section \ref{Section VII}.

\section{Related Work}
\label{Section II}
In PFL, some studies have been devoted to addressing the problem of FL under data heterogeneity from the perspective of knowledge forgetting. Shoham et al. \cite{shoham2019overcoming} considered knowledge forgetting in FL by analogy with the catastrophic forgetting of lifelong learning and related multi-task learning, and introduced a penalty term to the loss function to avoid forgetting previously learned knowledge. Xu et al. \cite{xu2022acceleration} reported the forgetting issue in local clients by empirically showing the increasing loss of previously learned data after the local training and alleviated knowledge forgetting by regularizing locally trained parameters with the loss on generated pseudo data. Unlike these studies, this study primarily employs KD and knowledge calibration techniques to mitigate the problem of forgetting.

Recent studies \cite{lee2022preservation, lu2023federated, he2022learning, yao2023f, huang2022learn,jin2022personalized, su2025dkd} introduced the KD technique into local training to address forgetting, which is relevant to our research. FedNTD \cite{lee2022preservation} introduced not-true distillation from the global model to prevent clients from forgetting global knowledge. Similarly, FedLMD \cite{lu2023federated} focused more on preserving minority label knowledge corresponding to the forgetting of each client. FedSSD \cite{he2022learning} applied mean squared error to build a distillation loss between the global model and the local model during local training, aiming to minimize the discrepancy between them. FedGKD \cite{yao2023f} averaged the global models cached on the server to create a historical global model, which is then used as a teacher model to guide the training of the client model. However, these studies aim to prevent the forgetting of global knowledge but do not take into account the forgetting of personalized knowledge. 

A few studies focus on the problem of personalized knowledge forgetting \cite{huang2022learn, jin2022personalized, su2025dkd}. Huang et al. \cite{huang2022learn} employed KD in local updates, where the distillation process between updated and pre-trained local models provides both inter- and intra-domain knowledge. However, pre-training personalized local models and performing distillation incurs non-negligible additional computational and storage overhead, which poses a significant burden for resource-constrained IIoT edge clients. pFedSD \cite{jin2022personalized} enables clients to distill the knowledge from previous personalized models into the current local models, thereby accelerating the recovery of personalized knowledge for newly initialized clients. Similarly, DKD-pFed \cite{su2025dkd} introduced the method of decoupled KD \cite{zhao2022decoupled} to distill the personalized model of the previous round. However, these studies are not concerned with the forgetting of global knowledge. In addition, Yao et al. \cite{yao2024rethinking} sought to preserve both global knowledge and local personalized knowledge simultaneously. After local training, an adaptive knowledge matrix is used to fuse knowledge from local, global, and historical models. The fused knowledge is then distilled into the local model. Although the purpose is the same, this study adopted the method of time dimensional progressive self-distillation and knowledge calibration.

Overall, existing studies primarily focus on the issues of global knowledge forgetting and local knowledge forgetting as separate research areas. They mainly employ KD  for knowledge extraction and retention but fail to adequately address the integration and preservation of both types of knowledge. Furthermore, most studies require the retention of teacher models, such as global models or previous local models, which imposes additional storage and computational burdens on resource-constrained IIoT edge devices. Lastly, in terms of KD specifics, the aforementioned studies adopt relatively traditional frameworks, where either the global model or the historical personalized model serves as the teacher to guide training across all local epochs. However, the global model lacks personalized knowledge, and the historical personalized model may not have fully converged, resulting in insufficient confidence on local data and the potential introduction of erroneous knowledge into the local update. Therefore, there is an urgent need to explore a new PFL method that achieves a balanced integration of global and personalized knowledge. 

\section{Motivation and Theoretical Analysis}
\label{Section III}
\subsection{Historical Personalized Knowledge Forgetting}
\subsubsection{Global model aggregation results in the erosion of historical personalized knowledge within local models}
The local model \( w_k^t \) of the client \( k \) is trained using its local dataset \( D_k \), allowing \( w_k^t \) to effectively capture the personalized characteristics of the local data, such as specific feature distributions and class biases. However, during global model aggregation (e.g., FedAvg), the personalized parameters of each client model may be overshadowed or diluted due to the weighted averaging process. Assuming that the parameters of each local model \( w_k^t \) can be decoupled into global parameters \( w_{\text{global}}^t \) and local personalized parameters \( w_{\text{local}, k}^t \) \cite{collins2021exploiting, sun2021partialfed}, the parameters of client $k$'s local model can be expressed as $ w_k^t = w_{\text{global}}^t + w_{\text{local}, k}^t $. After global model aggregation, the new global model parameters are computed as $w_{\text{global}}^{t+1} = \sum_{k=1}^{K} \frac{n_k}{n} \left( w_{\text{global}}^t + w_{\text{local}, k}^t \right)$ which can be further decomposed into $ w_{\text{global}}^{t+1} = w_{\text{global}}^t + \sum_{k=1}^{K} \frac{n_k}{n} w_{\text{local}, k}^t $, where \( n_k \) denotes the number of samples on client \( k \) and \( n = \sum_{k=1}^{K} n_k \) represents the total number of data samples across all $K$ clients. Since \( w_{\text{local}, k}^t \) is closely related to the local data distribution of client \( k \), the personalized parameters \( w_{\text{local}, k}^t \) often vary significantly across clients. During the averaging process in global aggregation, these differences are weakened or canceled out, resulting in the loss of local personalized knowledge in the global model. Furthermore, as the number of aggregation rounds increases, the problem of forgetting historical personalized knowledge becomes more pronounced, with the global model gradually losing the knowledge it learned from earlier local data.
\subsubsection{The non-uniform participation of clients exacerbates the issue of historical personalized knowledge forgetting} In real-world distributed federated learning environments, client devices may be unable to participate in every training round due to factors such as unstable network connections, limited device resources, or strict local privacy policies. Additionally, the server typically does not sample all clients in each training round but instead randomly selects a subset of clients for model updates. This non-uniform participation of clients can exacerbate the problem of local personalized knowledge forgetting. For example, when client \( k \) has not participated in training for an extended period, the global model may drift further away from capturing the personalized characteristics of client \( k \). As a result, when the client re-joins training, it must conduct additional local training to minimize the loss function \( \mathcal{L}_k(w_k) \) and recover its previously learned personalized knowledge.

\subsection{Local Personalization-Induced Global Generalization Knowledge Forgetting}
Global generalization knowledge refers to the features and patterns learned by the model from the global dataset that possess universality and generalizability. These features enable the model to make accurate predictions on data from different clients. However, as discussed in the previous section, in federated learning, each client starts training from the global model \( w_{g}^{t} \) in each new training round and performs local training on their data, which can be expressed as $w_k^{t+1} = w_{g}^{t} - \eta \nabla \mathcal{L}_k(w_{g}^{t}; D_k)$, where \( \nabla \mathcal{L}_k(w_g^{t}; D_k) \) is the gradient of the loss function, reflecting the features and patterns of the client’s data \( D_k \). If the client's data \( D_k \) differs significantly from the global dataset \( D \), the gradient direction may deviate substantially from the global optimal gradient direction \( \nabla \mathcal{L}(w_g^{t}; D) \). This gradient direction bias causes the client model to gradually optimize in a direction different from the global model during local updates. As the number of local training rounds increases, this bias accumulates, causing the client model to drift further away and potentially converge to a local minimum that better fits its local data distribution. This process leads to the forgetting of the generalizable features learned by the global model, akin to the ``catastrophic forgetting" issue in continual learning \cite{wang2024comprehensive}.

Overall, based on the theoretical analysis above, this study posits that achieving a balance between global knowledge and local personalized knowledge on the client side is critical for improving the convergence speed and accuracy of local personalized models.

\section{Methodology}
\label{Section IV}
\begin{figure*}[t]
\centering
\includegraphics[width=0.8\textwidth]{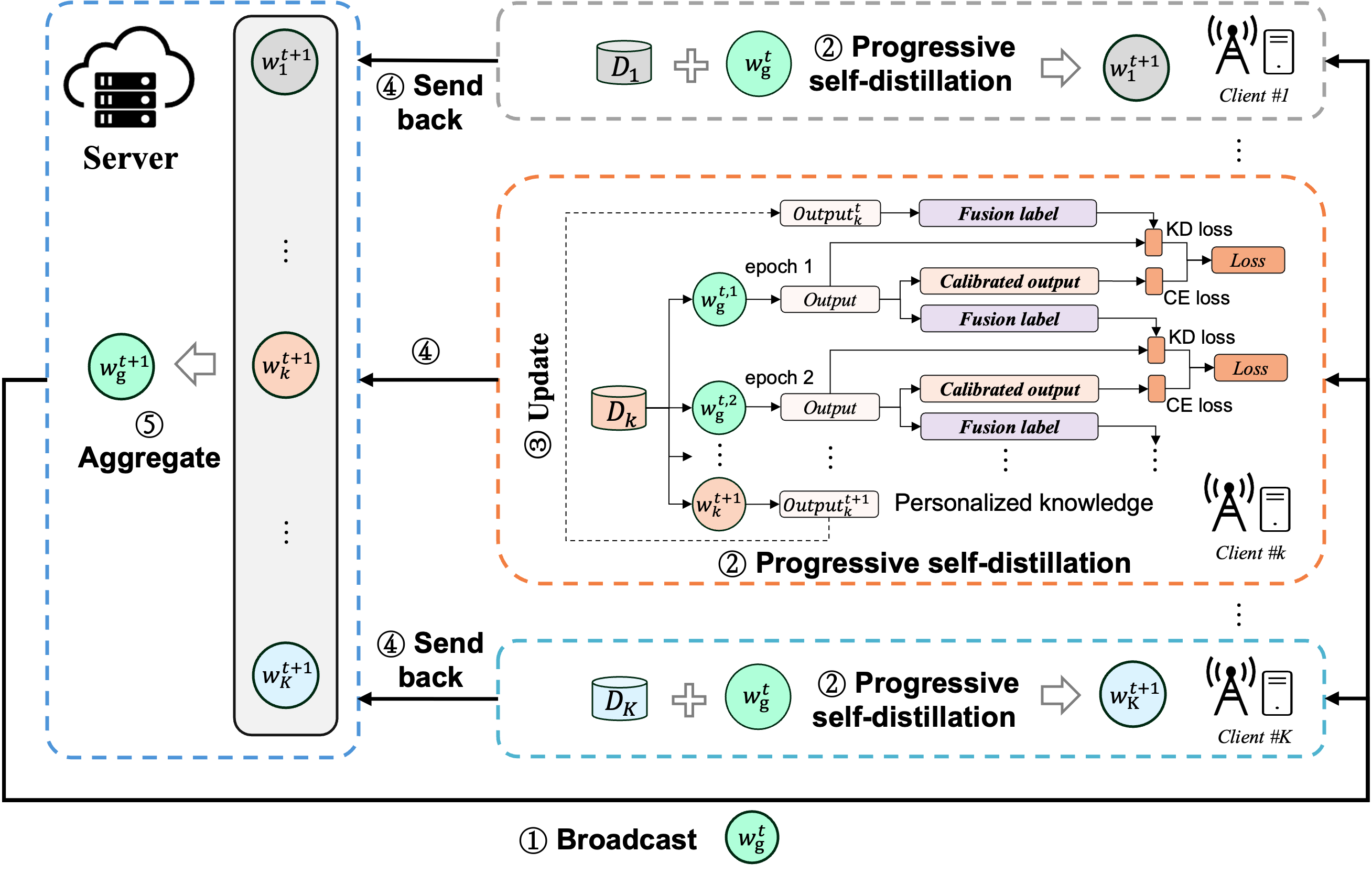}
\caption{The framework of federated progressive self-distillation with logits calibration.}
\label{fig1}
\end{figure*}
\subsection{Federated Progressive Self-Distillation Framework}
As illustrated in Figure \ref{fig1}, the modifications introduced by the FedPSD algorithm are concentrated on the client side. During each communication round, FedPSD retains the output of the locally trained personalized model on client \( k \) to guide self-distillation training in the subsequent round. Specifically, in communication round \( t \), when client \( k \) is selected to participate in training, the global model \( w_g^{t} \) provided by the server is used to initialize the local model. During the local self-distillation process, FedPSD assumes that the local model \( w_g^{t,e} \) at the \( e \)-th training epoch serves as the teacher model for the subsequent \( (e+1) \)-th epoch, where \( w_g^{t,(e+1)} \) is the student model. The output of \( w_g^{t,e} \) is fused with the ground truth labels to create smoother and more confident soft labels, which are then used to guide the training of the student model. Notably, for the initial epoch of student model training (\( e=1 \)), FedPSD uses a fusion of the ground truth labels and the historical personalized knowledge to construct the teacher's soft labels. This ensures the retention and recall of personalized knowledge during the initial training stage. Additionally, FedPSD employs calibrated logits to construct the local cross-entropy loss \(\mathcal{L}_{CE}\).

\subsection{Progressive Self-Distillation Loss}

\subsubsection{Reviewing Historical Personalized Knowledge}
To address the problem of forgetting historical personalized knowledge, this study preserves the output soft labels of the historical personalized model on client \( k \) to guide the initial epoch training of the current round's local model. This mechanism enables the local model to quickly recall its prior personalized knowledge. However, since the historical personalized model may not have fully converged, its output might contain erroneous knowledge. To mitigate this, the historical model's output is calibrated using the ground truth labels. Specifically, assuming that in the \((t-1)\)-th communication round of federated learning, the output probability vector of client \( k \)'s personalized model after local training is \( P_k^{t-1} \), and the corresponding ground truth label is \( Y_k \), the calibrated fused label \( H_k^{t-1} \) is defined as follows:

\begin{equation}
H_k^{t-1} = \alpha P_k^{t-1} + (1-\alpha)Y_k,
\end{equation}
where \( \alpha \in [0,1] \) is an adaptively adjusted weight. In the early stages of federated learning, the performance of the local client model tends to fluctuate due to insufficient convergence, necessitating a higher weight for the ground truth label \( Y_k \). As communication rounds progress and the historical model converges, greater weight should be assigned to the historical model's output \( P_k^{t-1} \). Thus, \( \alpha \in [0,1] \) should progressively increase with the number of communication rounds. A linear growth strategy is adopted in this study, with the calculation given by:

\begin{equation}
\alpha = \frac{t}{t_{\text{total}}},
\end{equation}
where \( t \) represents the current communication round, and \( t_{\text{total}} \) denotes the total number of communication rounds.
\subsubsection{Progressive Fusion of Personalized Knowledge and Global Knowledge}
To mitigate the issue of global knowledge forgetting during local training, this study introduces a progressive self-distillation strategy along the temporal dimension. This approach leverages the output knowledge of the model from the previous epoch to guide the training of the model in the subsequent epoch, thereby achieving a gradual integration of local personalized knowledge and global knowledge. However, since the local model may not have converged and lacks sufficient personalized knowledge at this stage, its output confidence is relatively low. Solely relying on the output of \( w_g^{t,e} \) to guide the training of \( w_g^{t,(e+1)} \) could result in the latter learning erroneous or ambiguous knowledge. To address this, as in the previous section, the output knowledge of the model from the previous epoch is calibrated using ground truth labels.

Specifically, assuming that in communication round \( t \), the predicted probability of client \( k \)'s model updated during the \((e-1)\)-th epoch is \( P_k^{t,e-1} \), and the ground truth label is \( Y_k \), the calibrated fused label \( H_k^{t,e-1} \) is computed as follows.
\begin{equation}
    \label{eq-5}
   H_k^{t,e-1}=\alpha P_k^{t,e-1}+(1-\alpha)Y_k
\end{equation}

Overall, the distillation loss for the \( e \)-th Epoch in the progressive self-distillation process is formulated as:  
\begin{equation}
    \label{eq-6}
    \mathcal{L}_{KD} = \mathcal{KL}(H_k^{t,e-1} \| P_k^{t,e}),
\end{equation}  
where \( \mathcal{KL}(\cdot \| \cdot) \) represents the Kullback-Leibler divergence, which measures the difference between two probability distributions. For \( e = 1 \), \( H_k^{t,e-1} = H_k^{t-1} \).

\subsection{Logits Calibration-Based Local Cross-Entropy Loss}
In federated learning, during the local training of client \( k \), the presence of non-independent and identically distributed (Non-IID) data often leads to bias in the local cross-entropy loss. Specifically, for client \( k \), assume its data distribution is \( P(x, y) = P(x \mid y)P(y) \). Given an input \( x \), the predicted label by the model is \( \hat{y} = \arg\max_y f_y(x) \), where \( f_y(x) \) represents the model's logits for class \( y \). During local training, softmax cross-entropy is typically used as the loss function to maximize the conditional probability \( P(y \mid x) \), indirectly minimizing the classification error rate \( P_{x, y}(y \neq \hat{y}) \). By Bayes' theorem, \( P(y \mid x) \propto P(x \mid y)P(y) \), which indicates that the conditional probability \( P(y \mid x) \) is jointly determined by the class-conditional distribution \( P(x \mid y) \) and the class prior \( P(y) \).  

However, in Non-IID environments, the class distributions across clients exhibit significant heterogeneity, typically reflected as skewed class priors \( P(y) \). Under such circumstances, training the model directly by maximizing \( P(y \mid x) \), although it minimizes the overall classification error \( P_{x, y}(y \neq \hat{y}) \), causes the model to favor locally frequent classes, often at the expense of accuracy on infrequent local classes. Thus, minimizing the overall error alone is insufficient to ensure fair classification performance across all classes in imbalanced class distributions. To address this issue, the class probabilities can be calibrated during inference to achieve a balanced form: $ P_{\text{Bal}}(y \mid x) \propto \frac{1}{L} P(x \mid y)$, where \( L \) is the total number of classes, and \( \frac{1}{L} \) represents a uniform prior distribution over all classes. This calibration effectively eliminates the influence of the class prior \( P(y) \), enabling the test results to approximate an estimation of \( P(x \mid y) \). The calibrated conditional probability can be expressed as: $ P_{\text{Bal}}(y \mid x) = \frac{P(y \mid x)}{P(y)}$. Therefore, the optimization objective based on the balanced probabilities is defined as minimizing the balanced classification error, which is as follows.  
	\begin{equation}
	\label{eq7}
	\arg\max_{y \in L} P_{\text{Bal}}(y \mid x) = \arg\max_{y \in L} \frac{P(y \mid x)}{P(y)}
	\end{equation}  

For softmax cross-entropy, where the class probability \( P(y \mid x) \propto e^{f_y(x)} \), substituting into the Eq. \ref{eq7} yields:  
\begin{equation}
\label{eq-8}
\arg\max_{y \in L} P_{\text{Bal}}(y \mid x) = \arg\max_{y \in L} (f_y(x) - \ln P(y)).
\end{equation}  

As shown in Equation \ref{eq-8}, subtracting the logarithm of the class prior from the logits \( f_y(x) \) will effectively adjust for the influence of \( P(y) \) during inference.

Therefore, this study calibrates the output probabilities during the training process of the client \(k\) in Non-IID environments to indirectly achieve an inference optimization objective equivalent to \(\arg\max_{y \in L} (f_y(x) - \ln P(y))\). Specifically, the local training cross-entropy loss is modified by adding the logarithm of the class prior weight \(\ln P(y)\) to the logits \(f_y(x)\), guiding the model to prioritize optimization for locally frequent classes. The calibrated output class probability is defined as follows.  
\begin{equation} 
\label{eq-9}
P_{\text{calibrated}}(y \mid x) = \frac{P(y)e^{f_y(x)}}{\sum_{y^{\prime} \in L} P(y^{\prime}) e^{f_{y^{\prime}}(x)}}
\end{equation} 
where \(P(y)\) denotes the prior distribution of class \(y\) in the local client’s data. Based on this calibrated probability, the calibrated local cross-entropy loss function is defined as follows.  
\begin{equation}
\label{eq10}
\mathcal{L}_{CE}(f(x), y) = -\log\frac{P(y)e^{f_y(x)}}{\sum_{y^{\prime} \in L} P(y^{\prime}) e^{f_{y^{\prime}}(x)}}.
\end{equation} 

In summary, the loss function for the local training process of client \(k\) is defined as follows. 

\begin{equation}
\label{eq11}
\mathcal{L}_{local} = \mathcal{L}_{CE} + \mathcal{L}_{KD},
\end{equation} 
where \(\mathcal{L}_{CE}\) represents the calibrated cross-entropy loss, and \(\mathcal{L}_{KD}\) denotes the knowledge distillation loss.

\section{Experiments and Discussion}
\label{Section V}
\subsection{Datasets and Data Partition}
\subsubsection{Datasets} To ensure a fair comparison, we selected three of the most commonly used datasets in the field of federated learning: MNIST \cite{lecun1998gradient}, CIFAR-10, and CIFAR-100 \cite{krizhevsky2009learning}.

\subsubsection{Data Partition} To simulate Non-IID data scenarios, we adopted two different data partitioning strategies: pathological sharding  \cite{mcmahan2017communication} and Latent Dirichlet Allocation (LDA) \cite{li2021model}.

\textbf{Sharding:} The pathological sharding strategy achieves non-IID data distribution by slicing the data based on labels, where each slice is referred to as a shard. The dataset is divided into multiple shards, which are then allocated to different clients. The degree of data heterogeneity is determined by the number of shards owned by each client, corresponding to the number of categories each client has. Specifically, suppose the dataset contains \( M \) training samples. We define the number of shards as \( S \) and divide the dataset into \( S \times K \) groups, where \( K \) is the number of clients, and each group contains \( M / (S \times K) \) samples. Each client then randomly selects \( S \) groups as its local dataset. This strategy considers only statistical heterogeneity, as all clients have datasets of equal size, and there is no overlap in data samples across different clients. In this study, for the MNIST dataset, we set \( S = 2 \); for the CIFAR-10 dataset, we set \( S = 2, 3, 5, 10 \); and for the CIFAR-100 dataset, we set \( S = 10 \).

\textbf{LDA:} The LDA strategy simulates the heterogeneity of client data distributions by modeling them based on a Dirichlet distribution. Using the Dirichlet distribution's concentration parameter \( \alpha \), training samples of different categories are assigned to each client in varying proportions to achieve non-IID distributions. The parameter \( \alpha \) determines the degree of heterogeneity in the data distribution: as \( \alpha \) increases, the heterogeneity decreases. Given a parameter \( \alpha \), we sample each client’s data distribution from Dir(\( \alpha \)), which specifies the proportion of each category in the client’s dataset. Based on these sampled proportions, data samples are randomly assigned to clients. Under this strategy, the size and category distribution of each client’s dataset varies, potentially including majority classes, minority classes, or even missing classes, which better reflect real-world scenarios. In this study, for the MNIST dataset, we set \( \alpha = 0.1 \); for the CIFAR-10 dataset, \( \alpha = \{0.05, 0.1, 0.3, 0.5\} \); and for the CIFAR-100 dataset, \( \alpha = 0.1 \).

\begin{figure*}[t]
\centering
\includegraphics[width=\textwidth]{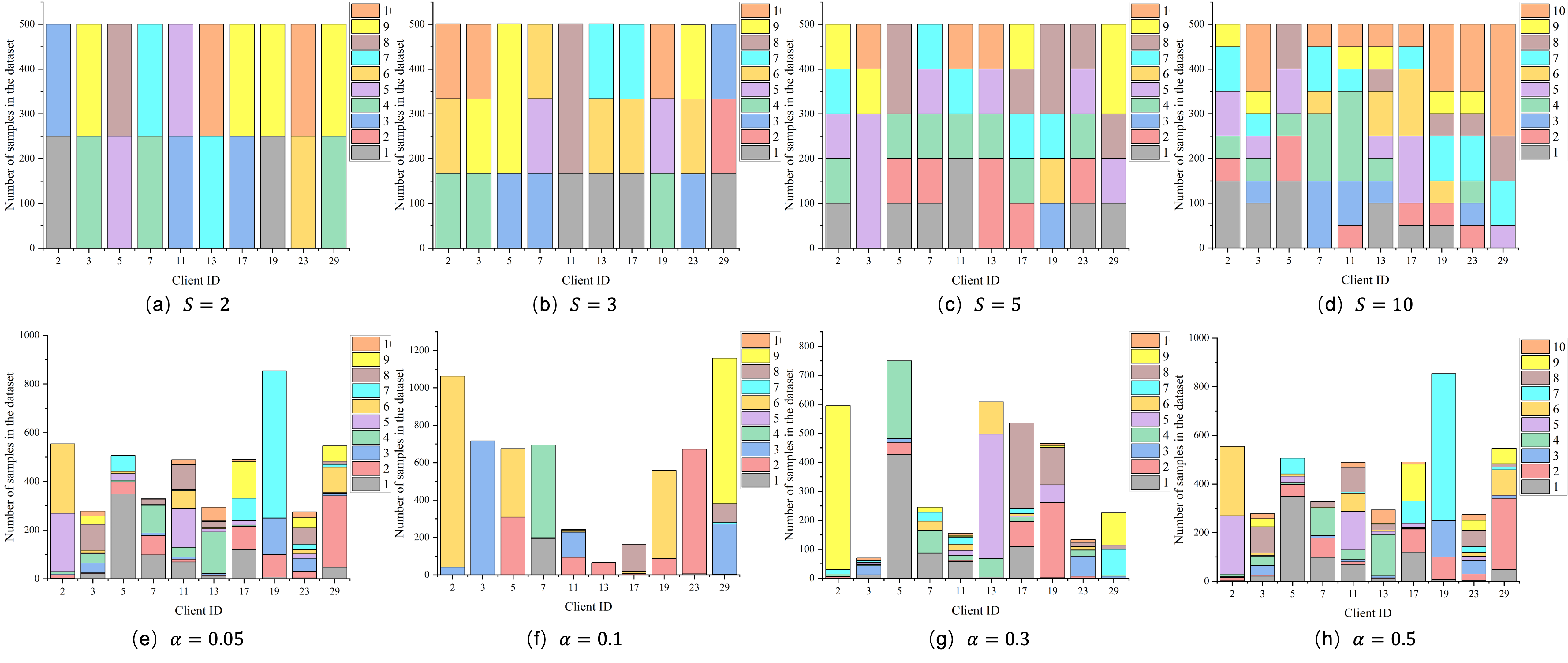}
\caption{Two distinct partitioning strategies were employed to segment the CIFAR-10 dataset, with visualizations of the distribution of different clients and their corresponding datasets. We randomly selected 10 clients for illustrative purposes.where $S$ denotes the dataset partitioned using the sharing pathological sharding strategy, and $\alpha$ represents the dataset partitioned using the LDA strategy.}
\label{fig3}
\end{figure*}

Figure \ref{fig3} presents the visualization of dataset distributions for 10 randomly selected clients on the CIFAR-10 dataset under two different data partitioning strategies.

\subsection{Baseline Methods}
We compared the proposed method with a range of widely recognized mainstream federated learning algorithms, including FedAvg\cite{mcmahan2017communication}, FedProx\cite{li2020federated}, FedCurv\cite{shoham2019overcoming}, FedNova\cite{wang2020tackling}, SCAFFOLD\cite{karimireddy2020scaffold}, MOON \cite{li2021model}, FedNTD \cite{lee2022preservation}, and FedLMD\cite{lu2023federated}, each of which has made significant contributions to the field of federated learning. 

\subsection{Implementation Details}
To ensure a fair comparison, we followed the experimental settings outlined in prior studies \cite{mcmahan2017communication, li2020federated} and fixed the training seed to maintain consistency in client selection and ensure the reproducibility of experiments. In all experiments, we employed a unified network architecture, SimpleCNN, consisting of two convolutional layers and two fully connected layers. In subsequent sections, we will discuss the performance of our method under different network architectures.  

Specifically, we set \( N = 100 \) total clients, with each client performing 5 local training epochs per communication round. The total number of communication rounds was set to 200, and in each round, 10\% of the clients (i.e., 10 clients) were randomly selected for training. For each client, training was conducted using the cross-entropy loss function and the SGD optimizer. The learning rate was set to 0.01 and decayed by a factor of 0.99 after each communication round. Weight decay was set to \( 1 \times 10^{-5} \), and the momentum for SGD was set to 0.9. The batch size was fixed at 50. In addition, we employed techniques such as random cropping, random horizontal flipping, and normalization for data augmentation, consistent with previous studies.

\subsection{Evaluation metrics}
To ensure reproducibility, the reported results are averaged over three independent runs with different random seeds. The performance of the algorithms is evaluated using two metrics: (1) the Top-1 test accuracy of the globally aggregated model on the server and (2) the average Top-1 test accuracy across all local clients which is defined as follows.
\begin{equation}
    \label{eq-em01}
    \overline{A}{_{t}}=\frac{1}{K}\sum_{i=1}^{K}A_{t,i}^{l}
\end{equation}
where, \( K \) denotes the total number of clients involved in the training process, and \( A_{t,i}^{l} \) represents the test Top-1 accuracy of the client \( i \) at communication round \( T \) (before aggregation) on its local test dataset.\textcolor{black}{Unless otherwise specified, the client accuracy mentioned in this paper refers to the average test accuracy of the clients. Correspondingly, the server accuracy denotes the test accuracy of the globally aggregated server-side model.}
\subsection{Experimental Results}
\subsubsection{Accuracy}

\begin{table*}
    \centering
    \caption{The average client Top-1 accuracy of different methods when using the pathological sharding strategy to partition the dataset.\textcolor{black}{The best accuracy is denoted by \textcolor{red}{red} values, and the second best by \textcolor{blue}{blue} value.}}
    \label{tab-1}
    \resizebox{0.65\textwidth}{!}{%
    \begin{tabular}{c|c|cccc|c} 
    \hline
    \multirow{2}{*}{Method} & MNIST & \multicolumn{4}{c|}{CIFAR-10} & CIFAR100  \\
                            & $S=2 $  & $S=2 $  & $S=3$   & $S=5 $  & $S=10 $ & $S=10$      \\ 
    \hline
    FedAvg                  & 55.92 & 20.45 & 28.71 & 38.64 & 55.77 & 8.15      \\ 
    \hline
    FedProx                 & 65.55 & 23.89 & 30.60 & 42.28 & 53.29 & 8.28      \\
    FedCurv                 & 56.05 & 24.08 & 28.47 & 38.15 & 52.10 & 7.99      \\
    FedNova                 & 59.82 & 21.84 & 30.28 & 42.94 & 54.65 & 10.52     \\
    SCAFFOLD                & 89.59 & 35.87 & 43.17 & 59.45 & 69.48 & 21.36     \\
    MOON                    & 55.23 & 20.10 & 28.39 & 41.38 & 56.59 & 8.17      \\
    FedNTD                  & 94.45 & \textcolor{blue}{42.66} & 53.26 & 64.38 & 70.51 & 27.80     \\
    FedLMD                  & \textcolor{blue}{95.12} & 42.42 & \textcolor{blue}{54.09} & \textcolor{blue}{65.23} & \textcolor{blue}{71.84} & \textcolor{blue}{28.69}     \\ 
    \hline
    Ours                    & \textcolor{red}{95.84} & \textcolor{red}{60.31} & \textcolor{red}{65.39} & \textcolor{red}{72.09} & \textcolor{red}{74.21} & \textcolor{red}{36.92}     \\
    \hline
    \end{tabular}
}
   
\end{table*}

\begin{table*}
    \centering
 
    \caption{The average client Top-1 accuracy of different methods when using the LDA strategy to partition the dataset.\textcolor{black}{The best accuracy is denoted by \textcolor{red}{red} values, and the second best by \textcolor{blue}{blue} value.}}
    \label{tab-2}
    \resizebox{0.7\textwidth}{!}{%
    \begin{tabular}{c|c|cccc|c} 
    \hline
    \multirow{2}{*}{Method} & MNIST & \multicolumn{4}{c|}{CIFAR-10}  & CIFAR100  \\
                            & $\alpha=0.1$ & $\alpha=0.05$ & $\alpha=0.1$ & $\alpha=0.3$ & $\alpha=0.5$ & $a=0.1$     \\ 
    \hline
    FedAvg                  & 85.47 & 23.74  & 34.88 & 46.22 & 54.75 & 19.00     \\ 
    \hline
    FedProx                 & 83.35 & 20.85  & 29.04 & 45.06 & 50.39 & 15.51     \\
    FedCurv                 & 85.80 & 24.54  & 33.34 & 44.42 & 51.41 & 18.26     \\
    FedNova                 & 71.36 & 10.00  & 24.40 & 44.08 & 52.90 & 15.53     \\
    SCAFFOLD                & 88.63 & 10.00  & 21.46 & 59.91 & 66.14 & 30.72     \\
    MOON                    & 84.43 & 22.38  & 34.93 & 48.30 & 55.89 & 19.17     \\
    FedNTD                  & 94.62 & 36.69  & 50.79 & 61.63 & 67.92 & 31.95     \\
    FedLMD                  & \textcolor{blue}{95.21} & \textcolor{blue}{38.19}  & \textcolor{blue}{51.51} & \textcolor{blue}{62.52} & \textcolor{blue}{68.66} & \textcolor{blue}{32.78}     \\ 
    \hline
    Ours                    & \textcolor{red}{96.20} & \textcolor{red}{50.23}  & \textcolor{red}{62.87} & \textcolor{red}{70.13} & \textcolor{red}{72.61} & \textcolor{red}{37.01}    \\
    \hline
    \end{tabular}
 }
    \end{table*}

\begin{table*}

    \centering
    \caption{The Top-1 accuracy of the global server model across different methods.\textcolor{black}{The best accuracy is denoted by \textcolor{red}{red} values, and the second best by \textcolor{blue}{blue} value.}} 
    \label{tab-3}
    \resizebox{0.8\textwidth}{!}{%
    \begin{tabular}{c|cc|cccc|cc} 
    \hline
    \multirow{2}{*}{Method} & \multicolumn{2}{c|}{MNIST} & \multicolumn{4}{c|}{CIFAR-10}  & \multicolumn{2}{c}{CIFAR100}  \\
                            & $S=2$   & $\alpha=0.1$              & $S=2$   & $S=5$   & $\alpha=0.05$ & $\alpha=0.1$ & $S=10$  & $\alpha=0.1$                 \\ 
    \hline
    FedAvg                  & 95.35 & 96.15              & 49.18 & 66.29 & 39.19  & 54.78 & 32.16 & 36.39                 \\
    \hline
    FedProx                 & 95.12 & 95.28              & 50.10 & 64.63 & 39.77  & 51.69 & 28.30 & 31.23                 \\
    FedCurv                 & 95.54 & 96.39              & 52.37 & 63.44 & 41.51  & 53.35 & 29.60 & 34.91                 \\
    FedNova                 & 93.25 & 86.75              & 44.92 & 63.65 & 10.00  & 35.91 & 29.81 & 33.25                 \\
    SCAFFOLD                & 96.90 & 94.20              & 57.79 & 73.27 & 10.00  & 26.68 & \textcolor{blue}{37.68} & \textcolor{red}{40.64}                 \\
    MOON                    & 95.87 & 96.03              & 49.18 & 65.62 & 39.07  & 54.65 & 32.11 & 36.37                 \\
    FedNTD                  & 96.70 & 96.61              & 63.49 & 73.09 & 47.29  & 63.81 & 36.67 & 38.62                 \\
    FedLMD                  & \textcolor{blue}{97.02} & \textcolor{blue}{96.96}              & \textcolor{blue}{63.54} & \textcolor{blue}{74.12} & \textcolor{blue}{49.31}  & \textcolor{blue}{63.76} & 36.81 & 38.34                 \\ 
    \hline
    Ours                    & \textcolor{red}{97.41} & \textcolor{red}{96.98}              & \textcolor{red}{64.71} & \textcolor{red}{74.25} & \textcolor{red}{51.44}  & \textcolor{red}{65.74} & \textcolor{red}{39.45} & \textcolor{blue}{38.87}                 \\
    \hline
    \end{tabular}
    }
    \end{table*}
    
We conducted a comprehensive evaluation of the proposed method and compared it with several mainstream approaches. As shown in Tables \ref{tab-1}-\ref{tab-3}, the experimental results demonstrate that the proposed method outperforms the baseline methods in terms of Top-1 accuracy on both client and server sides across multiple datasets, thereby validating its effectiveness. Specifically, the proposed method exhibits superior performance in both the average model accuracy on clients and the global model accuracy on the server. For the average model accuracy on clients, our method achieves a 32.1\% improvement on average compared to baseline methods and a 9.6\% improvement over the state-of-the-art (SOTA) methods when using the pathological sharding strategy to partition the CIFAR-10 dataset. Notably, under the dataset partitioning strategy with \( S=2 \), our method achieves a maximum improvement of 39.8\%. Significant improvements were also observed on the MNIST and CIFAR-100 datasets. When using the Latent Dirichlet Allocation (LDA) strategy to partition the CIFAR-10 dataset, our method similarly demonstrated outstanding performance, achieving an average improvement of 24\% over baseline methods and 8.7\% over SOTA methods. Even under extremely imbalanced scenarios, such as when \( \alpha = 0.05 \), the proposed method provides effective enhancements. Additionally, significant improvements were observed on the MNIST and CIFAR-100 datasets. These comparative results highlight the effectiveness of our strategy.  
  
In addition, as shown in Table \ref{Section III}, for the global model Top-1 accuracy on the server, our method also demonstrated excellent performance. When applied to the CIFAR-10 dataset with various data partitioning strategies, it achieved an average accuracy improvement of 11.68\% over baseline methods. On the CIFAR-100 dataset, the proposed method similarly achieved significant improvements, surpassing current SOTA methods. These findings indicate that our method is effective in preserving local client knowledge. 

\subsubsection{Communication Efficiency}
\begin{figure*}[t]
\centering
\includegraphics[width=\textwidth]{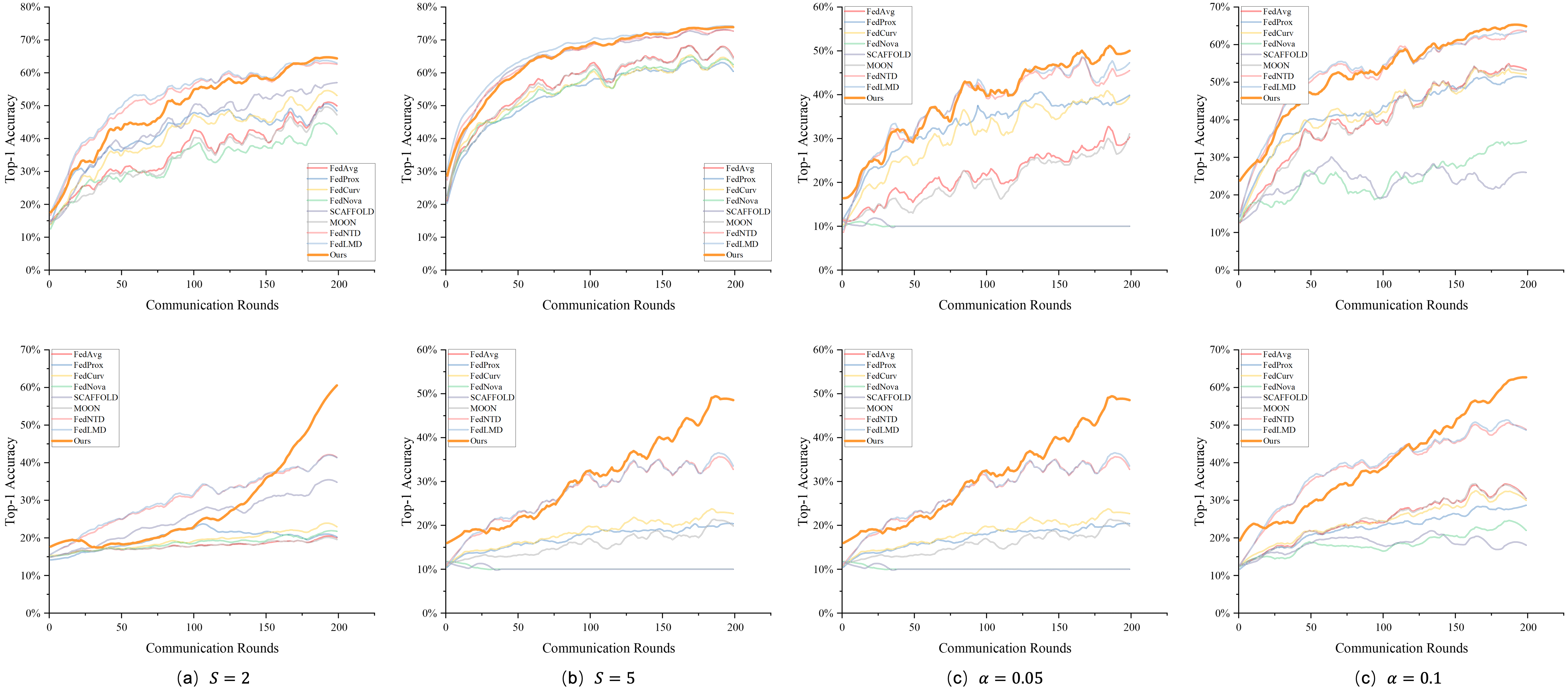}
\caption{Top-1 accuracy during training for different methods, where $S$ denotes the dataset partitioned using the sharing pathological sharding strategy, and $\alpha$ represents the dataset partitioned using the LDA strategy.The first row of figures illustrates the accuracy of the globally aggregated model during the training process, whereas the second row depicts the average accuracy of the clients throughout the training}
\label{fig2}
\end{figure*}

We also place particular emphasis on the execution efficiency of federated learning algorithms, a critical factor for resource-constrained real-world applications. To evaluate efficiency, we compare the number of communication rounds required by different methods to achieve the same final accuracy as FedAvg. Figure \ref{fig2} illustrates the variation in client and server accuracy after each communication round across different methods. Tables \ref{tab-4} and \ref{tab-5} provide detailed statistics on the number of communication rounds required by each method to reach the final accuracy of FedAvg.

\begin{table}
    \centering
    \caption{The number of communication rounds required by different methods to achieve the same final accuracy as FedAvg on the average client side of CIFAR-10 dataset when applying various partitioning strategies (N/A indicates that the accuracy does not reach the target even after 200 epochs).\textcolor{black}{The best number of communication rounds is denoted by \textcolor{red}{red} values, and the second best by \textcolor{blue}{blue} value.}} 
    \label{tab-4}
    \begin{tabular}{c|cc|cc} 
    \hline
    \multirow{3}{*}{Methods} & \multicolumn{4}{c}{CIFAR10}                              \\ 
    \cline{2-5}
                             & \multicolumn{2}{c|}{Sharding} & \multicolumn{2}{c}{LDA}  \\
                             & $S=2$ & $S=5$                     & $\alpha=0.05$ & $\alpha=0.1$           \\ 
    \hline
    FedAvg                   & 200 & 200                     & 200    & 200             \\ 
    \hline
    FedProx                  & 69  & 92                      & 184    & N/A             \\
    FedCurv                  & 105 & 165                     & 128    & 148             \\
    FedNova                  & 106 & 94                      & N/A    & N/A             \\
    SCAFFOLD                 & 41  & 35                      & N/A    & N/A             \\
    MOON                     & 158 & 123                     & 184    & 146             \\
    FedNTD                   & \textcolor{blue}{21}  & \textcolor{blue}{24}                      & 49     & \textcolor{blue}{44}             \\
    FedLMD                   & \textcolor{blue}{21}  & \textcolor{red}{21}                      & \textcolor{red}{38}     & \textcolor{red}{41}              \\ 
    \hline
    Ours                     & \textcolor{red}{10}  & 59                      & \textcolor{blue}{46}     & 62              \\
    \hline
    \end{tabular}
    \end{table}
    
    \begin{table}
    
    \centering
    \caption{The number of communication rounds required by different methods to achieve the same final accuracy as FedAvg on the global server side for the CIFAR-10 dataset under various partitioning strategies (N/A indicates that the accuracy does not reach the target even after 200 epochs).\textcolor{black}{The best number of communication rounds is denoted by \textcolor{red}{red} values, and the second best by \textcolor{blue}{blue} value.}} 
    \label{tab-5}
    \begin{tabular}{c|cc|cc} 
    \hline
    \multirow{3}{*}{Methods} & \multicolumn{4}{c}{CIFAR10}                              \\ 
    \cline{2-5}
                             & \multicolumn{2}{c|}{Sharding} & \multicolumn{2}{c}{LDA}  \\
                             & $S=2$ & $S=5$                     & $\alpha=0.05$ & $\alpha=0.1$           \\ 
    \hline
    FedAvg                   & 200 & 200                     & 200    & 200             \\ 
    \hline
    FedProx                  & 100 & N/A                     & 89     & 173             \\
    FedCurv                  & 97  & 165                     & 85     & 156             \\
    FedNova                  & 167 & 168                     & N/A    & N/A             \\
    SCAFFOLD                 & 82  & 65                      & N/A    & N/A             \\
    MOON                     & 143 & 123                     & N/A    & 136             \\
    FedNTD                   & 46  & 66                      & 58     & \textcolor{red}{50}              \\
    FedLMD                   & \textcolor{red}{40}  & \textcolor{blue}{64}                      & \textcolor{blue}{39}     & \textcolor{red}{50}             \\ 
    \hline
    Ours                     & \textcolor{blue}{44}  & \textcolor{red}{63}                      & \textcolor{red}{38}     & \textcolor{blue}{62}              \\
    \hline
    \end{tabular}

    \end{table}

\textcolor{black}{The results demonstrate that our proposed method is highly competitive in terms of convergence speed and accuracy compared to other approaches.} On average, our method requires only 45 communication rounds for local clients to reach the accuracy of FedAvg after 200 rounds. Similarly, at the server level, our method achieves the same accuracy in just 52 communication rounds on average. Specifically, on the CIFAR-10 dataset with various partitioning strategies, our method excels in the Sharding strategy with $S=2$, requiring only 10 communication rounds to match FedAvg's 200-round accuracy for local clients. Additionally, under the LDA strategy with $\alpha=0.05$, our method demonstrates outstanding performance, achieving FedAvg's 200-round accuracy in just 46 communication rounds.

These comparative results highlight the significant advantage of our approach in reducing communication requirements. Furthermore, as data heterogeneity increases, our method exhibits a more pronounced advantage in balancing communication efficiency and accuracy. This indicates its robustness and adaptability to varying levels of data heterogeneity, further underscoring its effectiveness in addressing the challenges of non-uniform data distributions.

\section{Analysis and Discussion}
\label{Section VI}
To comprehensively understand and validate the effectiveness of our proposed method, this section first evaluates the importance and contribution of each component through ablation studies. Subsequently, we examine the performance of the method under various network architectures, different numbers of local training epochs, and varying client participation rates. Unless otherwise specified, all experimental settings are consistent with those described in the previous section.

\subsection{Ablation Study}
By evaluating each component individually, we clarified their roles and contributions in addressing the issue of knowledge forgetting. The experiments were conducted using the CIFAR-10 dataset with two distinct data partitioning strategies to comprehensively assess the effects of the components. Specifically, we performed ablation studies on the strategy of Reviewing Historical Personalized Knowledge (RHPK), Progressive Self-Distillation (PSD), and Calibrated Logit-based Loss (CLL).

As shown in Table \ref{tab-6}, the results demonstrate that employing the GSD component alone significantly improves the average accuracy of local models. Compared to the baseline, it achieves improvements of 16.88\% and 8.45\% under the two different partitioning strategies, respectively. Introducing the PSD component further enhances the average accuracy, with additional gains of 10.91\% and 4.04\%, respectively, over the baseline. This indicates that revisiting historical personalized knowledge effectively mitigates knowledge forgetting in local clients. Finally, incorporating the RELogits component leads to further performance gains of 12.38\% and 18.86\%, respectively. This demonstrates that adjusting the weight of different classes based on the local dataset distribution alleviates the problem of local models excessively fitting their own datasets while neglecting the collaborative benefits and knowledge sharing provided by the global model. In addition, when all three components are employed together, both the average accuracy of the local models and the accuracy of the global model are further enhanced, confirming the effectiveness of our method in addressing the challenge of local knowledge forgetting.

\begin{table}
    \centering
    \caption{The impact of different components on model performance ($\Delta \textcolor{green}{\uparrow}$ denotes the Average client Top-1 accuracy improvement compared to the baseline method).} 
    \label{tab-6}
    \resizebox{\columnwidth}{!}{
    \begin{tabular}{c|ccccc} 
    \hline
    DataSets                                                                   & RHPK & PSD & CLL                 & Top-1 Acc & $\Delta$ \\ 
    \hline
    \multirow{4}{*}{\begin{tabular}[c]{@{}c@{}}CIFAR10\\(S=2)\end{tabular}}    & \multicolumn{3}{c}{FedAvg(Baseline)} & 20.14     & N/A      \\ 
    \cline{2-6}
    & \checkmark   &              &              & 37.02     & 16.88 \textcolor{green}{$\uparrow$}   \\
    & \checkmark   & \checkmark   &              & 47.94     & 27.79 \textcolor{green}{$\uparrow$}    \\
    & \checkmark   & \checkmark   & \checkmark   & 60.31     & 40.17 \textcolor{green}{$\uparrow$}    \\ 
    \hline
    \multirow{4}{*}{\begin{tabular}[c]{@{}c@{}}CIFAR10\\($\alpha $=0.05)\end{tabular}} & \multicolumn{3}{c}{FedAvg(Baseline)} & 21.78     & N/A      \\ 
    \cline{2-6}
   & \checkmark   &              &             & 27.33     & 5.55  \textcolor{green}{$\uparrow$}    \\
   & \checkmark   & \checkmark   &             & 31.37     & 9.59  \textcolor{green}{$\uparrow$}    \\
   & \checkmark   & \checkmark   & \checkmark  & 50.23     & 28.45 \textcolor{green}{$\uparrow$}    \\
   \hline
    \end{tabular}
    }
    \end{table}

\subsection{Impact of Different Network Architectures}
We evaluated the applicability of the proposed method across various network architectures. Using the CIFAR-10 dataset, we conducted additional experiments comparing the performance of two representative architectures, MobileNet and ResNet. The results are detailed in Table \ref{tab-7}. Experimental findings demonstrate that our method consistently achieves stable performance and significant improvements across multiple network architectures. These results highlight the method's applicability and robustness in real-world federated learning scenarios involving diverse backbone architectures.

\begin{table}
    \centering
    \caption{Comparison of Average client Top-1 accuracy across different network architectures.\textcolor{black}{The best accuracy is denoted by \textcolor{red}{red} values, and the second best by \textcolor{blue}{blue} value.}}
    \label{tab-7}
    \resizebox{0.42\textwidth}{!}{
    \begin{tabular}{c|ccc} 
    \hline
    Method        & Simple CNN     & ResNet 10      & VGG 11          \\ 
    \hline
    FedAvg        & 20.45          & 18.05          & 17.84           \\
    FedProx       & 23.89          & 17.46          & 16.93           \\
    FedCurv       & 24.08          & 18.18          & 17.59           \\
    FedNTD        & \textcolor{blue}{42.66}          & \textcolor{blue}{44.44}          & 31.93           \\
    FedLMD        & 42.42          & 32.28          & \textcolor{blue}{32.72}           \\ 
    \hline
    \textbf{Ours} & \textcolor{red}{60.31} & \textcolor{red}{54.12} & \textcolor{red}{36.40}  \\
    \hline
    \end{tabular}
    }
    \end{table}

\subsection{Impact of Local Training Epochs}  
\begin{figure}[t]
\centering
\includegraphics[width=0.4\textwidth]{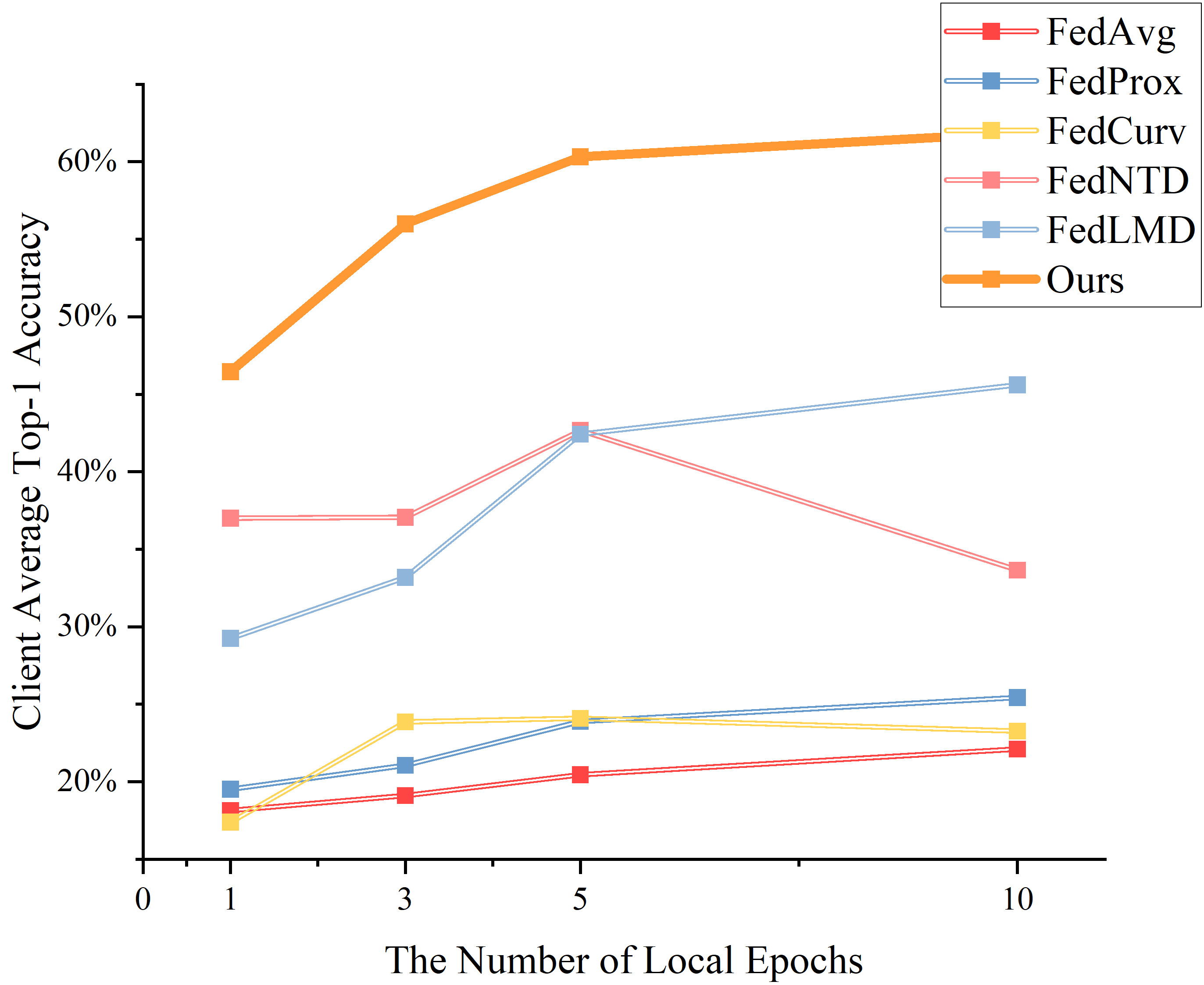}
\caption{Average client Top-1 accuracy at different epoch configurations.}
\label{fig-3}
\end{figure}

We examined the impact of varying the number of local training epochs per communication round on model accuracy. In addition to the fixed configuration of $Epoch = 5$, we evaluated scenarios with $Epoch = 1, 3, 10,$ and $20$. The experimental results, presented in Figure \ref{fig-3}, demonstrate that the proposed method consistently delivers stable and substantial performance improvements across all configurations. Notably, as $Epoch$ increases, methods such as FedNTD exhibit a decline in accuracy during the later stages of training, primarily due to overfitting to the local data distribution induced by larger $Epoch$ values. In contrast, the proposed method sustains consistent and progressive performance gains, even with higher $Epoch$ values, and consistently surpasses the performance of all other methods. These results highlight the superior adaptability and robustness of our approach.

\subsection{Impact of the Number of Participating Clients}  

In federated learning, a key issue worth exploring is the number of clients participating in each communication round. In real-world applications, variations in communication efficiency between individual nodes and the central server often result in fluctuating client participation per communication round. To simulate and emphasize this phenomenon observed in practical federated learning scenarios, we experimented with different client participation rates. Specifically, the proportion of clients selected in each communication round was set to 5

\begin{figure}[htp]
\centering
\includegraphics[width=0.4\textwidth]{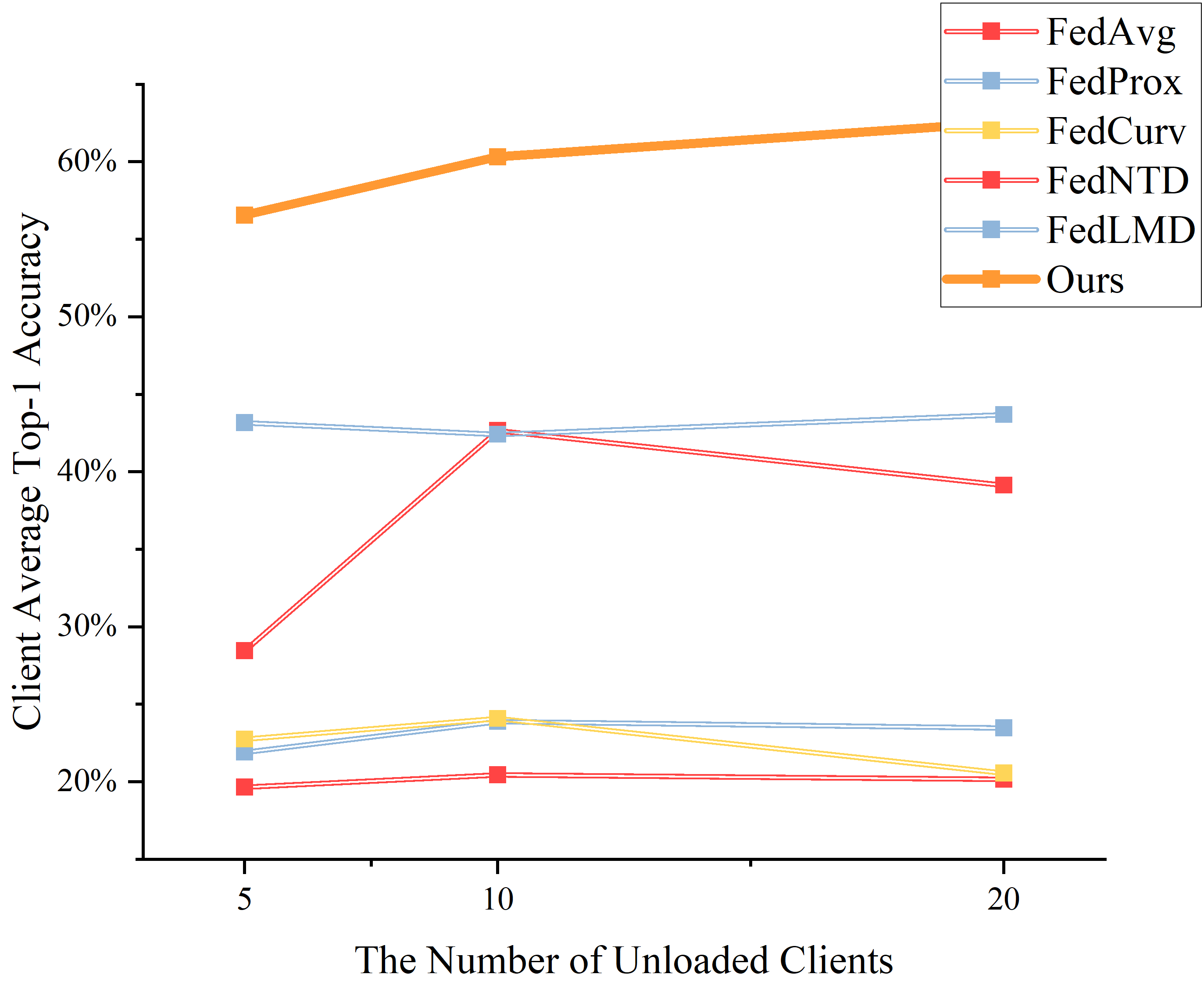}
\caption{Average client Top-1 accuracy at different clients configurations.}
\label{fig-4}
\end{figure}

As shown in Figure \ref{fig-4}, the accuracy of various methods improves as the number of participating clients increases. However, our proposed method consistently achieves superior performance even when the client participation rate is low. This finding highlights the capability of our approach to effectively retain local knowledge while also assimilating additional knowledge from other clients, even under scenarios with limited client participation in communications.

\section{Conclusion}
\label{Section VII}

This study proposes Federated Progressive Self-Distillation (FedPSD), a novel approach designed to address the challenges of global knowledge retention and local personalized knowledge forgetting in personalized federated learning. FedPSD incorporates a client-side knowledge distillation mechanism, augmented by calibrated local training losses, to achieve an optimal balance between global generalization knowledge and local personalized knowledge. Specifically, FedPSD preserves the personalized model outputs generated during local training at each communication round, leveraging them to guide self-distillation in subsequent rounds and facilitating the efficient recall of historical personalized knowledge.  Moreover, FedPSD introduces a progressive self-distillation loss and a logits-calibrated local training loss, both aimed at alleviating catastrophic forgetting of global knowledge. Experimental evaluations demonstrate that FedPSD consistently improves both the convergence speed and the accuracy of client-side models while simultaneously enhancing the performance of the global model.  

FedPSD provides an effective federated learning strategy for personalized IIoT edge intelligence, improving the personalization capabilities of models while safeguarding user privacy. Future work will focus on extending the applicability of FedPSD to more diverse scenarios and refining the algorithm to address the challenges posed by increasingly complex distributed learning environments.

\bibliography{IEEE_REF}{} 
\bibliographystyle{IEEEtran}

\vspace{11pt}






\vfill

\end{document}